\documentclass{article}

\usepackage{arxiv}

\usepackage{caption}
\usepackage{subcaption}
\usepackage{xcolor}
\usepackage{authblk}
\usepackage{hyperref}
\usepackage{multirow}
\usepackage{listings}
\usepackage{soul}
\usepackage{grffile}
\usepackage{float}
\usepackage{fullpage}
\usepackage{booktabs}
\usepackage{array}
\usepackage{chngpage}
\usepackage{amsmath}
\usepackage{amssymb}
\usepackage{graphicx}
\graphicspath{ {../images/} }

\newcommand{\nummethods}{20}
\newcommand{\ringxorsum}{RING+XOR+SUM}
\newcommand{\ringxor}{RING+XOR}
\newcommand{\ring}{RING}
\newcommand{\xor}{XOR}

\newcommand{\supplfigringxorsum}{4}

\title{How good Neural Networks interpretation methods really are? A quantitative benchmark.}

\author{
Antoine Passemiers\textsuperscript{1,\textasteriskcentered}, Pietro Folco\textsuperscript{2,\textasteriskcentered}, Daniele Raimondi\textsuperscript{1,\textdagger}, Giovanni Birolo\textsuperscript{2}, Yves Moreau\textsuperscript{1}, Piero Fariselli\textsuperscript{2}
}

\begin{document}
\renewcommand{\thefootnote}{\fnsymbol{footnote}}
\maketitle
\footnotetext{\textsuperscript{1}ESAT-STADIUS, KU Leuven, Leuven, Belgium}
\footnotetext{\textsuperscript{2}Department of Medical Sciences, University of Torino, Torino, Italy}
\footnotetext{\textsuperscript{*}These authors contributed equally to this work.}
\footnotetext{\textsuperscript{\textdagger}Corresponding author: daniele.raimondi@kuleuven.be}
\begin{abstract}
Saliency Maps (SMs) have been extensively used to interpret deep learning models decision by highlighting the features deemed relevant by the model. They are used on highly nonlinear problems, where linear feature selection (FS) methods fail at highlighting relevant explanatory variables. However, the reliability of gradient-based feature attribution methods such as SM has mostly been only qualitatively (visually) assessed, and quantitative benchmarks are currently missing, partially due to the lack of a definite ground truth on image data. 
Concerned about the apophenic biases introduced by visual assessment of these methods, in this paper we propose a synthetic quantitative benchmark for Neural Networks (NNs) interpretation methods.
For this purpose, we built synthetic datasets with nonlinearly separable classes and increasing number of \emph{decoy} (random) features, illustrating the challenge of FS in high-dimensional settings. We also compare these methods to conventional approaches such as mRMR or Random Forests. 
Our results show that our simple synthetic datasets are sufficient to challenge most of the benchmarked methods. TreeShap, mRMR and LassoNet are the best performing FS methods. We also show that, when quantifying the relevance of a few non linearly-entangled predictive features diluted in a large number of irrelevant noisy variables, neural network-based FS and interpretation methods are still far from being reliable.
\end{abstract}


\section{Introduction}

Decision processes are different in machines and humans. The training of Neural Networks (NNs) is solely guided by the minimizing of a loss function, which incidentally and only partially aligns with human understanding of the problem, since the optimization of the loss function relies on the labeling of training samples rather than articulate expert insights, eventually making these models diverge from human expectations. An extreme example of such divergence are \textit{Clever Hans} predictors, which exclusively rely on dataset artifacts \cite{lapuschkin2019unmasking}. More often, their decision processes are influenced by contextual features which are not necessary causal, and which are thus acting as confounders (e.g. presence of wooden hurdles in horse pictures). Therefore, \textit{Clever Hans} predictors can appear proficient on training data but remain prone to potentially high generalization error in unseen settings or on independent test data (e.g. absence of wooden hurdles in horse pictures).

For these reasons, interpretability has become an emerging crucial aspect of Machine Learning (ML), and solutions for Explainable Artificial Intelligence (XAI) \cite{lapuschkin2019unmasking} are now also ethical requirements posed by institutions like the European Union \cite{holzinger2018current}. In many \emph{real-life} applications, a clear understanding of the model decision-making process is indeed necessary for its adoption in a production environment.
Interpretability is even more relevant in the context of nonlinear techniques with strong prediction power and modeling capabilities such as Deep Neural Networks (DNN) \cite{lapuschkin2019unmasking}, since linear models are notoriously easier to explain and can therefore be complemented with simple but sound and effective FS methods (LASSO \cite{kim2007interior}, shrunken centroid method \cite{tibshirani2002diagnosis}) that exploits the additivity of input variables~\cite{ESTbook}. 

Due to the intrinsic human-readability properties of simpler models (i.e. linear models, decision trees), they are often preferred in critical settings. On the other hand, complex nonlinear models such as DNNs exploit non-trivial interactions between input features, making them a model-of-choice for solving difficult tasks that require higher abstraction (e.g. Genome interpretation, protein folding and human-level real-time control \cite{galiana, alphafold, atari, dnc}). In theory, DNNs can ignore irrelevant features during the training phase, but this depends on the optimizer and the loss function landscape, which can be highly non-convex.

In recent years, gradient-based \emph{a posteriori} interpretation methods for DNNs such as Saliency Maps (SMs) rapidly gained popularity \cite{simonyan2013deep}. The most common approaches include Integrated Gradients \cite{DBLP:journals/corr/SundararajanTY17}, DeepLift \cite{DBLP:journals/corr/ShrikumarGK17}, Input $\times$ Gradient \cite{DBLP:journals/corr/ShrikumarGSK16}, SmoothGrad \cite{DBLP:journals/corr/SmilkovTKVW17} and Guided Backpropagation \cite{springenberg2014striving}. They can be easily applied to any NN model with libraries such as Captum \cite{captum2019github}. 

These methods have been developed mainly in the context of computer vision, and their ability to identify the \emph{salient} pixels for the classification of images was primarily qualitatively assessed by visually inspecting the obtained SMs in relation to the input images, because there was no \emph{ground truth} for what should be considered salient, since this concept is intrinsically model-dependent. To the best of our knowledge, quantitative benchmarks of the reliability of SM interpretations is currently missing, especially on non-image data. At the same time, several studies showing puzzling SM interpretation results and perplexing behaviors have been already published, raising concerns about these methods \cite{SMprob1, SMprob2, SMprob3}.

The interpretation of non-linear model is an extremely active field of research and, besides \emph{a posteriori} interpretation methods such as SMs, in recent years several DNNs architectures incorporating FS capabilities have been developed, such as CancelOut \cite{borisov2019cancelout}, DeepPINK \cite{lu2018deeppink}, LassoNet \cite{JMLR:v22:20-848},  FSNet \cite{singh2020fsnet}, Concrete Autoencoder \cite{abid2019concrete} and Diet-Net \cite{Romero17-iclr}. 

In this paper, we benchmarked the most recent nonlinear FS and interpretation methods for DNNs on synthetic, simplistic, yet challenging artificial datasets, comparing them with older and more conventional FS approaches. Each of our datasets consists of few (2-7) variables that jointly and nonlinearly correlate with the output class, as well as a variable number of irrelevant random features (decoys). To allow fair assessment, both relevant and irrelevant variables have been sampled with the exact same variance: each feature follows a uniform marginal distribution in 4 out of 5 of our datasets. In our fifth dataset, features have been simply standardized. In this way, methods that exploit variance signatures, such as Principal Component Analysis are not applicable. These datasets have been constructed in such a way that the classes cannot be segregated by linear decision boundaries, making linear FS methods totally unsuitable for the problem at hand. The synthetic nature of the datasets grants us complete knowledge of the predictive signal and allows us to quantitatively benchmark the ability of nonlinear FS methods to detect nonlinearly and jointly relevant features in controlled sample-to-feature ratio experimental settings. As baseline methods, we used additional FS methods that are not based on NNs, like Random forests (RFs) or minimum redundancy maximum relevance (mRMR).

Our results show that the DNN-based FS methods tested are not able to extract relevant features if they are diluted in a random set of noisy variables. 
Conversely, feature relevances extracted from RFs are more reliable on average. We obtained similar results while benchmarking several SM a posteriori interpretation methods. 
These results indicate that the field of FS and DNNs interpretation needs to further refine the available methods, and suggest that a standardized \emph{quantitative validation} of the newly proposed methods should be used to assess their performance and limits, to give users a more realistic idea of the situations in which these approaches are actually reliable.

\section{Methods}
\subsection{Benchmark datasets}

\subsubsection{Uniformly-distributed variables}

\begin{figure}
    \centering
    \includegraphics[width=\textwidth]{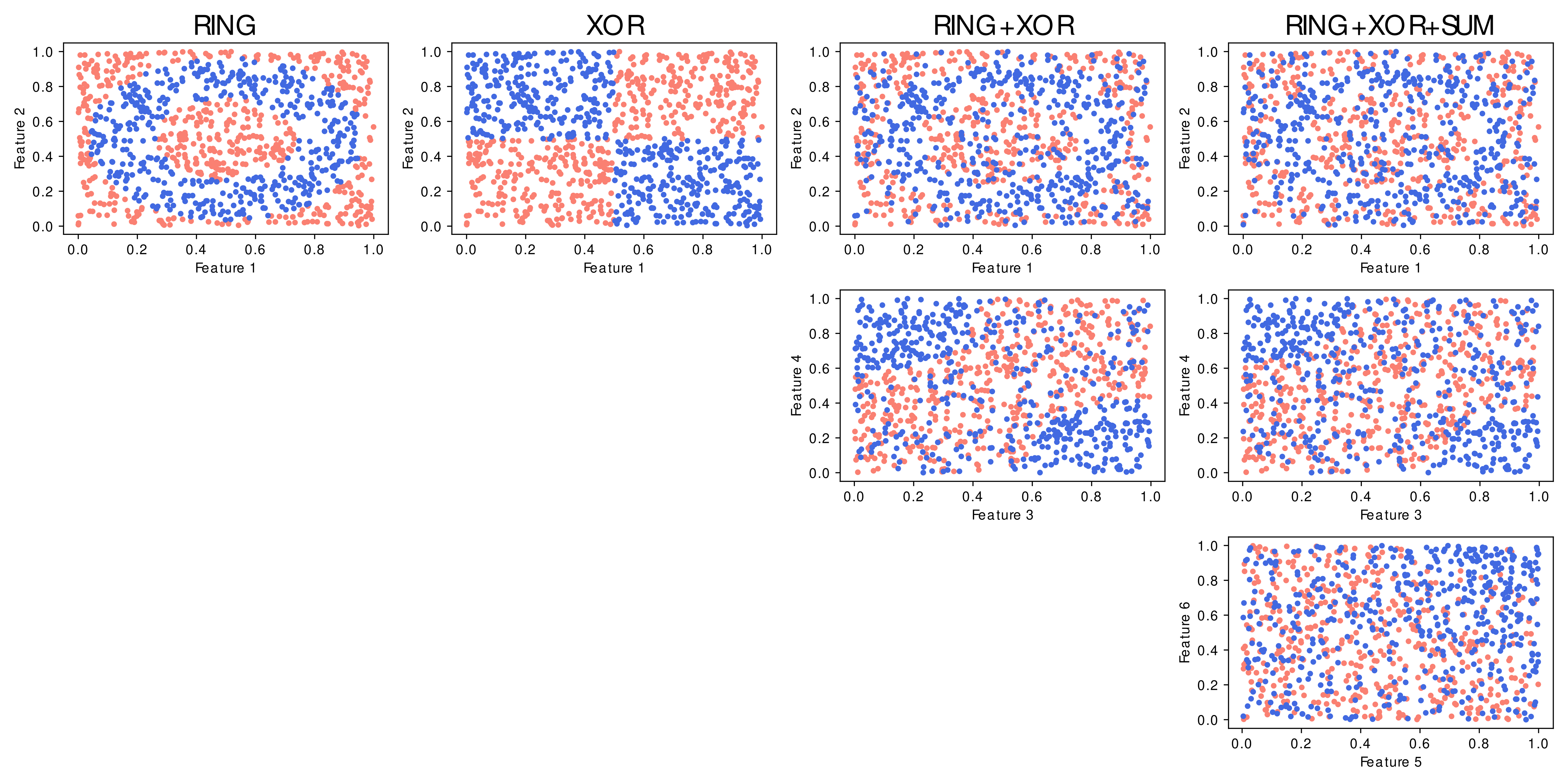}
    \caption{The four datasets, with their predictive features shown by pairs. Orange and blue correspond to the positive and negative classes, respectively. Each column is associated with one dataset, and each row corresponds to a distinct features pair.}
    \label{fig:outcomes}
\end{figure}

We first built four datasets representing archetypal nonlinear binary classification tasks. 
Each dataset contains $n=1000$ observations and $m=p+k$ features, uniformly distributed in the $[0,1]$ interval. $p$ and $k$ denote the number of predictive and irrelevant features, respectively. 
Each dataset was then built by attributing a label
to data points according to a nonlinear combination of the predictive features.
All the remaining features are effectively random with respect to the labels, and thus act as decoys when it comes to feature selection. 
The number of positives is equal to the number of negatives, to prevent any artifact due to class imbalance.
Here we describe the different characteristics of each outcome (label). They are also visually shown in Fig. \ref{fig:outcomes}.

\begin{itemize}
\item \ring: positive labels are associated to the points that form a bi-dimensional ring, defined by the features in positions $j \in \{0,1\}$. 
The total number of predictive features is 2. Points where assigned to the positive class when:
\[
\lvert \sqrt{(x_0 - 0.5)^2 + (x_1 - 0.5)^2} - 0.35 \rvert \leq 0.1151
\]

\item \xor: the bi-dimensional space formed by the features in position $j \in \{0,1\}$ is divided into 4 identical regions. Points in the upper left and lower right quadrants are labelled as positive samples. 
The total number of predictive features is 2. Points were considered positive when:
\[
(x_0 - 0.5)(0.5 - x_1) \ge 0
\]

\item \ringxor: the samples that are either positive samples in the \ring \ dataset (considering the features in position $j \in \{0,1\}$) or positive samples in the \xor \ dataset (considering the features in positions $j \in \{2,3\}$) are positive. This dataset thus contains 4 predictive features, in positions $j \in \{0,1,2,3\}$. 
Points were considered positives when they satisfied any of the following:
\begin{align*}
\lvert \sqrt{(x_0 - 0.5)^2 + (x_1 - 0.5)^2} - 0.35 \rvert 
\leq 0.0704
\\
(x_2 - 0.5)(0.5 - x_3) \ge 0.0337
\end{align*}

\item \ringxorsum: the samples that are positive in the RING, XOR  datasets or such that $x_4 + x_5 + \epsilon > 0.5$ are positive samples. $x_j$ denotes the $j$-th feature and $\epsilon$ is Gaussian noise sampled from $\mathcal{N}(\mu=0,\sigma=0.2)$. Like in the \ring \ and \xor \ datasets, the corresponding predictive features $\{0,1,2,3\}$ do not contain noise. In \ringxorsum \ the predictive features are in positions $\{0,1,2,3,4,5\}$. See Suppl. Fig. \supplfigringxorsum \ for a visual explanation.
Points were considered positives when they satisfied at least one of the following inequalities:
\begin{align*}
\lvert \sqrt{(x_0 - 0.5)^2 + (x_1 - 0.5)^2} - 0.35 \rvert 
\leq 0.0479
\\
(x_2 - 0.5)(0.5 - x_3) \ge 0.0598
\\
x_4 + x_5 + \mathcal{N}(\mu=0,\sigma=0.2) \ge
1.4074
\end{align*}
\end{itemize}
The four outcomes are shown in Fig.\ref{fig:outcomes}.

\subsubsection{Gaussian graphical model}

While for sufficiently large values of $m$ the datasets defined above are already challenging, we built an additional dataset (called DAG) characterized also by confounding effects. Indeed, these effects occur in most real-life settings, and are very likely to \emph{misguide} the different models towards learning predictive patterns from irrelevant associations. This dataset has been generated by a directed Gaussian graphical model. The exact procedure is detailed in Suppl. Mat. 1.1.
By construction, features can be categorized based on their degree of relevance:
\begin{itemize}
    \item Features $X_i$ that are highly relevant as they are causal for the target variable $Y$: $X_i \rightarrow \dotsc \rightarrow Y$
    \item Features $X_i$ that are weakly relevant as they correlate with the target variable only due to indirect effects, like in forks: $X_i \leftarrow \dotsc \leftarrow X_j \rightarrow \dotsc \rightarrow Y$
    \item Irrelevant features
\end{itemize}

\subsection{Benchmark procedure}

We assessed the reliability of FS methods on nonlinear ML tasks with a growing degree of difficulty by incrementally diluting the relevant features in the uniformly-distributed datasets. An exponential increase of the number $k$ of decoy features is added in each run: $m \in \{2, 4\} \cup K$ for XOR and RING datasets, $m \in \{4\} \cup K$ for RING+XOR and $m \in \{6\} \cup K$ for RING+XOR+SUM, where $K = \{8, 16, 32, 64, 128, 256, 512, 1024, 2048\}$. 
For each run, we set the number of samples to $n = 1000$.

For each run and dataset, we assessed the performance of both predictors and FS methods. First, for each embedded FS method that relies on a predictive model, we evaluated the latter using a 6-fold cross-validation procedure. Then we computed the AUROC and AUPRC of each step accordingly. We also reported these metrics for a baseline NN without prior or posterior FS. In the paper, we simply refer to it as ``Neural Network''.

Second, we evaluated the ability of each FS algorithm to rank the predictive features higher than the decoys. More specifically, we quantified the latter as the percentage of predictive features among the highest $p$ and $2p$ top-ranked features, where $p$ corresponds to the number of truly predictive features in each dataset ($p=2$ for XOR and RING, $p=4$ for RING+XOR, $p=6$ for RING+XOR+SUM, $p=7$ or $81$ in DAG depending on the definition used). We refer to them as best $p$ and best $2p$ scores for short.
To ensure fair comparison, and avoid assigning good performance to badly-designed FS methods (where feature importances are influenced by their position/indices in the data matrix), we randomly permuted the columns of the input data matrix in each fold of the $k$-fold cross-validation procedure. For feature attribution methods (e.g. saliency maps), only the points from the held-out sets of the 6-fold cross-validation have been used to select features.

\subsection{Machine Learning models}
To ensure a fair comparison between NN-based FS methods, we tried to reuse the same neural architecture whenever possible. By default, NNs have been implemented with Pytorch \cite{pytorch}. The data has been centered by replacing each input vector $x_i$ by $2 x_i - 1$, so each feature ranges between $-1$ and $1$. The default model is a three-layer perceptron with LeakyReLU activation functions, with a 0.2 slope for negative values. The 2 hidden layers have $16$ neurons each. Additionally, L2 regularisation on the parameters is used, with a regularisation parameter of $10^{-2}$. The model was trained with the Adam optimizer~\cite{kingma2014adam} for up to $1000$ epochs, with a learning rate of $0.005$ and a batch size of $64$. A scheduler guided the optimisation of each model by adjusting the learning rate over time, decreasing it by a $0.9$ factor every time the total loss function stagnates for 10 epochs (with a cooldown of 5 epochs). Gaussian noise, with a standard deviation of 0.05, was added to the inputs in order to regularise the models further.
In order to minimize overfitting risks and get the best out of each NN-based approach, we implemented an early stopping criterion. We saved the NN parameters at each epoch and kept the ones that minimize the loss function on a held-out set composed of 20\% of the training data. The training was interrupted prematurely if the validation loss had not decreased during the last 5 epochs.

All feature attribution methods are based on the architecture described above. However, due to the peculiarities of some embedded FS methods or the constraints of their implementation, these latter methods might build on subtle variants of this architecture. When relevant, these differences are explained in the next section.

\subsection{Feature selection methods}

The benchmarked algorithms belong to three categories: feature attribution, embedded, and filter methods. We now describe them in details.
\subsubsection{Feature attribution methods}
Feature attribution methods propose an \emph{a posteriori} interpretation of the method $M$ by approximately reconstructing the decision process followed by $M$ in order to produce the prediction $y_i$. 
Gradient-based interpretation methods for NNs like Saliency Maps (SMs) belong to this category. Given a trained NN model $M$, the forward pass $M(x_i)$ of sample $x_i$ is computed, alongside with the gradient $\partial M(x_i) /\partial x_i$ of the target output $y_i = M(x_i)$ with respect to the input $x_i$. The gradient values identify which positions in $x_i$ are the most relevant for the prediction. Indeed, because the gradient points towards the direction of steepest descent, the highest components of the gradient indicate which input variables require the least change to produce the largest variation in the output $y_i$. In this study we used the Captum \cite{captum2019github} library to benchmark the Integrated Gradients \cite{DBLP:journals/corr/SundararajanTY17}, Saliency \cite{simonyan2013deep}, DeepLift \cite{DBLP:journals/corr/ShrikumarGK17}, Input $\times$ Gradient \cite{DBLP:journals/corr/ShrikumarGSK16}, SmoothGrad \cite{DBLP:journals/corr/SmilkovTKVW17}, Guided Backpropagation \cite{springenberg2014striving}. From the same library, we also benchmarked non-gradient-based interpretation methods like Deconvolution \cite{mahendran2016salient}, Feature Ablation, Feature Permutation \cite{molnar2020interpretable} and Shapley Value \cite{castro2009polynomial} interpretation approaches.
For SmoothGrad, data was injected with random noise sampled from a zero-centered Gaussian distribution with 0.1 standard deviation (Captum's implementation of \texttt{NoiseTunnel}). The operation has been repeated $50$ times per input vector. For the Integrated Gradients, DeepLift, Feature Ablation and Shapley Value Sampling methods, the $0$ vector was supplied as baseline point. The baseline point is used for different purposes depending on the method. For example, the basepoint provided for the Integrated Gradients method defines the point from which to compute the integral and smooth the feature attribution vector.
Because all these gradient-based methods only provide instance-level feature importances, we computed the overall feature importances as the average of absolute values of the instance-level importances. Because it was \textit{a priori} unclear to us whether these instance-level scores should be computed on the training set or the validation set (with their difference explained by the generalization error), we compared the two settings in section \ref{attr-train-test}.

\subsubsection{Embedded FS methods}
The second category of FS approaches are the embedded methods, which operate \emph{at training time}, such as FSNet \cite{singh2020fsnet}, Concrete Autoencoder \cite{abid2019concrete}, CancelOut \cite{borisov2019cancelout}, Random Forest \cite{breiman2001random}, DeepPINK \cite{lu2018deeppink} and LassoNet \cite{JMLR:v22:20-848}. 
In particular, FSNet and Concrete Autoencoder jointly \emph{rank} features and train the model.

FSNet's architecture is composed of a selector and an encoder network, which branches into a classifier and a final sub-network comprising a decoder and a reconstruction layer. The selector consists in a matrix multiplication operation, involving a low-rank matrix of shape $k \times 2p$ obtained from a weights predictor, followed by a softmax activation (after addition of Gumbel noise to the logits): $2p$ features are selected, and we computed both best $p$ and best $2p$ scores using the feature importances as proposed in \cite{singh2020fsnet}. The weights predictor of the selector is composed of a fully-connected layer with no activation. We chose the identity function as activation for both the encoder and decoder, as no further refinement of the input features was deemed necessary. The reconstruction layer is the reverse operation of the selector network, and performs a matrix multiplication analogously. The corresponding low-rank matrix is predicted from a predictor module composed of a fully-connected layer with no activation. The classifier has the same architecture as described in the previous section, except that its input size is restricted to $2p$. The whole model was trained for $2000$ epochs. $30$ bins (latent size of $10$) were used to compute the inputs frequencies necessary to predict the low-rank matrices.

CancelOut is composed of the common architecture described in the previous section (trained in the same manner), preceded by a CancelOut layer. We experimented with 2 variants of the method:
1) with a Sigmoid activation and CancelOut weights regularisation (we denote the corresponding model by CancelOut Sigmoid for short), and 
2) with a Softmax activation layer and without regularization (CancelOut Softmax). In the former case, we used a $\lambda_1 = 0.2$ coefficient for the variance and a $\lambda_2 = 0.1$ for the regularisation term. Because L1 regularisation encourages importances weights to converge to 0.5 (sigmoid(0) = 0.5), we replaced it by the sum of CancelOut weights, unlike the original implementation. Indeed, this choice of regularisation better promotes sparsity among feature importances.
In both cases, CancelOut weights were initialised with the same value $\beta = 1$. The model has been trained for $300$ epochs, as the convergence of CancelOut weights requires longer time than the optimisation of the classifier alone.

All models built for evaluating the LassoNet \cite{JMLR:v22:20-848} feature selection method contained $32$ hidden neurons and ReLU activation functions, as only the latter were available among activation functions in the \texttt{lassonet} Python package.

To evaluate the Concrete Autoencoder (CAE), we used the \\ \texttt{concrete-autoencoder} Python package \cite{abid2019concrete}, and implemented the underlying classifier with Keras \cite{chollet2015keras} in accordance with the architecture described in the previous section. The CAE has been trained for 300 epochs, with initial temperature $10$ and final temperature $0.01$. The CAE has been trained twice, once for selecting $p$ and once for selecting $2p$ features. The model has been trained for $10$ epochs.
Each hidden layer is composed of $32$ neurons, and followed by a 20\% dropout.

DeepPINK requires knockoff features, that we generated in two different ways depending on the nature of the dataset. For the DAG dataset, we relied on the framework of Model-X Knockoff features designed by Candès et al. \cite{osti_10049247}. Details of Model-X Knockoff features are explained in Suppl. Mat. 1.2. For the remaining 4 datasets, knockoff features were generated by simply sampling a uniform distribution (the marginal distribution of the original data matrix $X$ is multivariate but uniform).

Random forests were grown with the Scikit-learn Python package \cite{scikit-learn} and are composed of $500$ trees each. We used the default parameters. We selected features from trained random forests using the widely-used impurity-based feature importance scores, as implemented in Scikit-Learn. More specifically, the importance of a feature is defined as the total reduction in Gini impurity caused by all node splits involving that feature, averaged across all trees in the forest.


\subsubsection{Filter methods}
The last category of methods that we considered is the filter-method algorithms, such as Relief \cite{rosario2015relief} or minimum redundancy maximum relevance (mRMR) \cite{peng2005feature}. These methods do not assign predictions; therefore no AUROC or AUPRC values are reported.
mRMR relies on statistical measures such as mutual information \cite{kraskov2004estimating}, which requires a discretisation of the input variables. Therefore, we divided each feature into 20 equally-sized bins, which we are sufficiently thin for capturing the nonlinear dependencies, and sufficiently large for robust estimation ($\sim$ 40 observations per bin).
AUROC and AUPRC were computed for FS methods based on a predictive model, and reported as a function of the input feature size.

\section{Results}

Feature Selection (FS) is widely used across many fields of science \cite{fs1,fs2,limitations}. The goal of FS algorithms is to identify or rank features in function of the \emph{predictive signal} they carry with respect to a prediction label. In this study we benchmarked \nummethods \ FS methods on 5 synthetic datasets (RING, XOR, RING+XOR, RING+XOR+SUM and DAG) providing nonlinear binary classification tasks (see Methods). These datasets represent classical ML problem, such as the XOR problem, the discrimination of points lying on a ring-shaped subspace and combinations thereof (see Methods for more details). 

\subsection{Random forests outperform other methods on \ring \ by a large margin}
In the \ring \ dataset, positive labels are associated to the points lying on a bi-dimensional ring defined by features in positions ${0,1}$ (see Suppl. Fig. S1).

\begin{figure}
    \includegraphics[width=\textwidth]{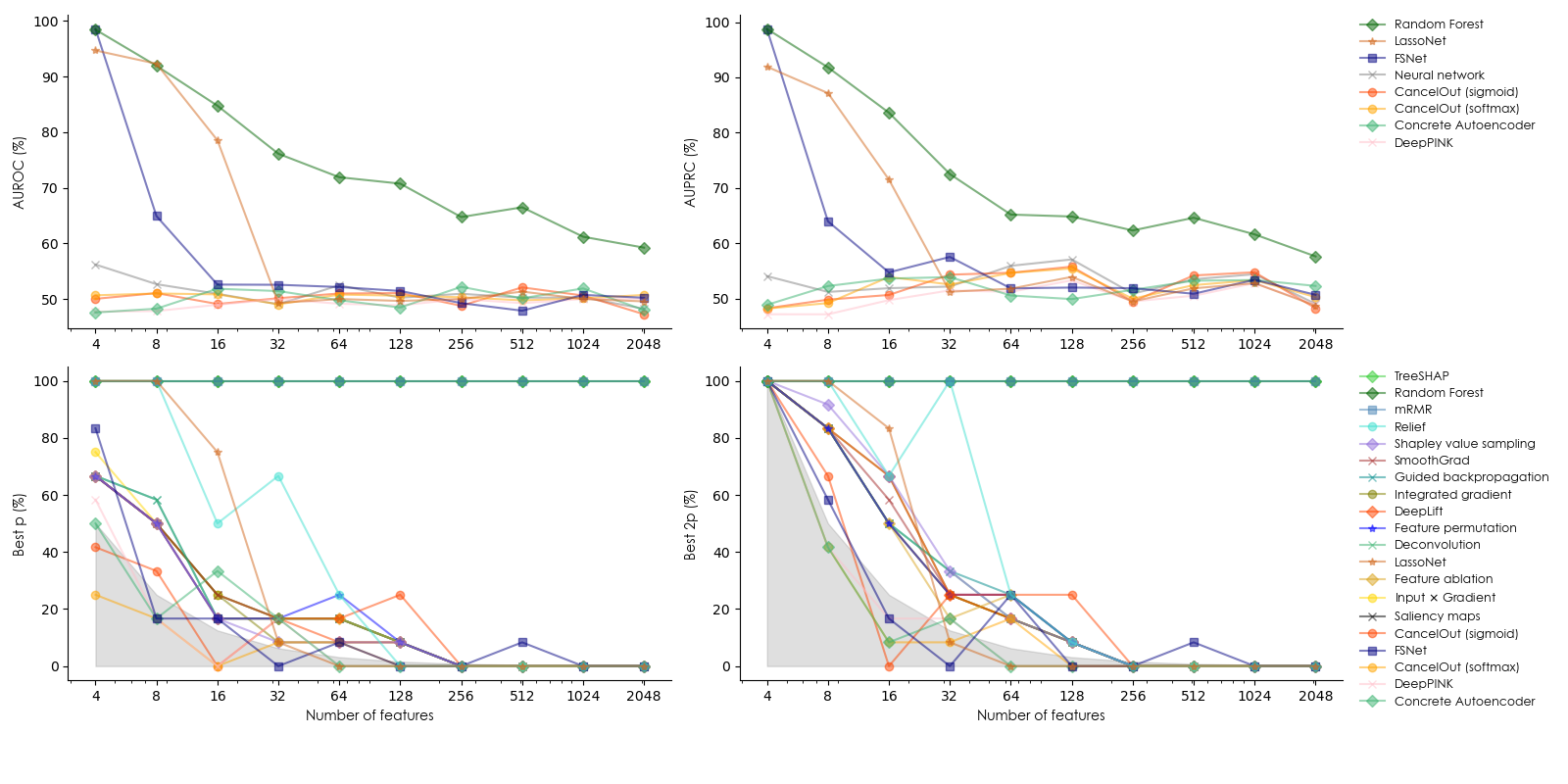}
    \caption{Performance of the different models and feature selection methods on the \ring \ dataset. (Top) AUROC and AUPRC of each trained model as a function of the number of features. (Bottom) Percentage of relevant features selected by each FS method in the top \(p\) and \(2p\), respectively. Shaded areas correspond to the best $p$ and best $2p$ scores of a dummy FS method that performs worse than random. Methods have been sorted by decreasing order of average performance in the legend.}
    \label{fig:benchmark-ring}
\end{figure}

Top panels in Fig. \ref{fig:benchmark-ring} show the AUROC and AUPRC of all trained models, as a function of the total number of features $m = p + k$. It is noteworthy that all models but the Random Forest have AUROCs and AUPRCs approaching a random predictor (50\%) for values of \(m\) greater than 32. This effect is consistent with the percentages of relevant features reported in the bottom graphs of panel Fig.  \ref{fig:benchmark-ring}.
Both Random forests and TreeSHAP perfectly re-identified the relevant features, even when $m = 2048$. However, their decaying performance (as $m$ increases) suggest that tree-based models may lose their FS capabilities in extremely high-dimensional ($m \gg 2048$).

\subsection{Neural networks are better suited for solving the \xor \ problem}

\begin{figure}
    \includegraphics[width=\textwidth]{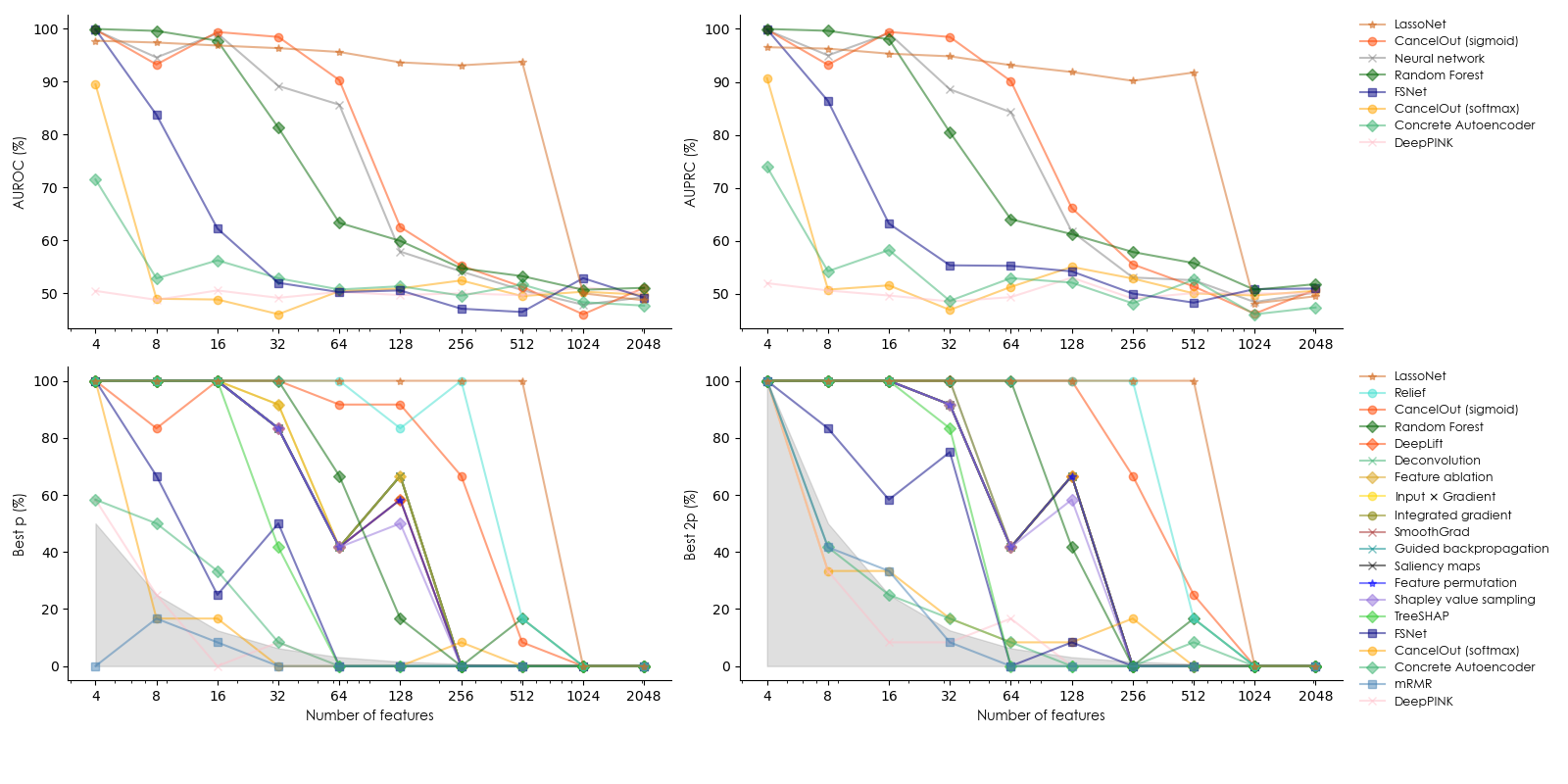}
    \caption{Performance of the different models and feature selection methods on the \xor \ dataset. (Top) AUROC and AUPRC of each trained model as a function of the number of features. (Bottom) Percentage of relevant features selected by each FS method in the top \(p\) and \(2p\), respectively. Shaded areas correspond to the best $p$ and best $2p$ scores of a dummy FS method that performs worse than random. Methods have been sorted by decreasing order of average performance in the legend.}
    \label{fig:benchmark-xor}
\end{figure}

As shown in Fig. \ref{fig:benchmark-xor}, NN models obtained overall better results, and mRMR underperformed, even for a small number of irrelevant features $k$.
In particular, LassoNet reached maximal best $p$ and best $2p$ scores for each $m$, and $> 90\%$ AUROC and AUPRC for each $m \le 512$. This relatively high predictive performance of LassoNet can be attributed to the internal cross-validation procedure used to find the optimal degree of sparsity of the NN parameters. The FS methods, besides LassoNet, that found the highest proportions of relevant features are Relief, CancelOut Sigmoid, and feature attribution methods like DeepLift or Deconvolution. Most methods undergo drastic performance losses when $m \ge 128$.

\subsection{RF, mRMR and LassoNet are the best performing methods on \ringxor}

\begin{figure}
    \includegraphics[width=\textwidth]{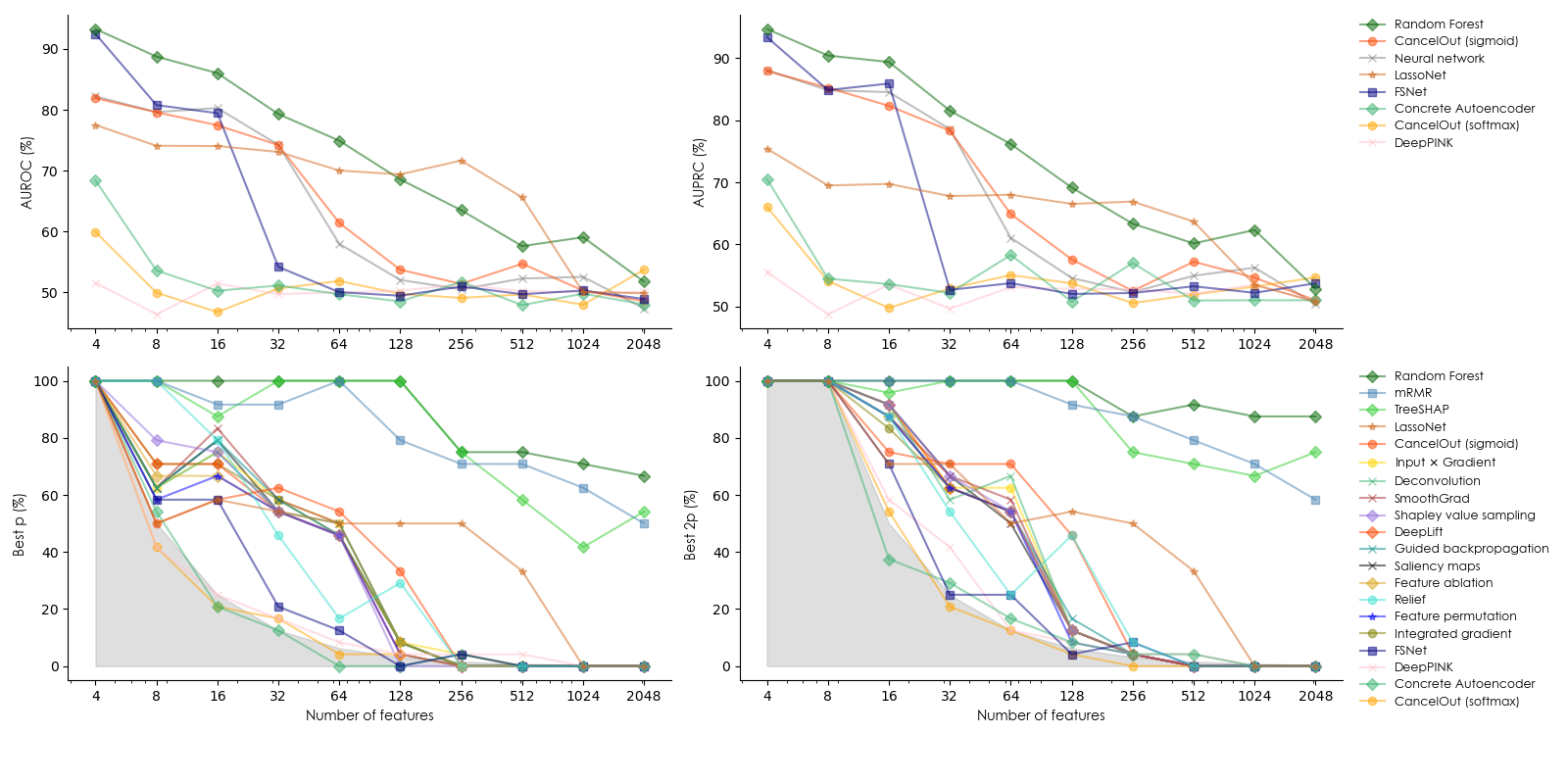}
    \caption{Performance of the different models and feature selection methods on the \ringxor \ dataset. (Top) AUROC and AUPRC of each trained model as a function of the number of features. (Bottom) Percentage of relevant features selected by each FS method in the top \(p\) and \(2p\), respectively. Shaded areas correspond to the best $p$ and best $2p$ scores of a dummy FS method that performs worse than random. Methods have been sorted by decreasing order of average performance in the legend.}
    \label{fig:benchmark-ring-xor}
\end{figure}

This dataset comprises four relevant features ($p = 4$). Similarly to what we observed for the \ring \ dataset, we see from Fig. \ref{fig:benchmark-ring-xor} and Tab. \ref{tab:benchmark} that Random Forests and TreeSHAP (which interprets learned Random Forests) perform the best, along with mRMR. Random Forests achieve high AUROC and AUPRC values and correctly rank the features in most settings. In particular, it constitutes the only FS method capable of ranking over the \( 60\% \) of relevant features among the top 4 features for \(m \le 2048\).
For any other method to achieve a similar percentage of relevant features selected $>$ \( 80\% \), the total number of features needs to be decreased to \(m = 16\), thus requiring to make the problem orders of magnitude simpler. Among these remaining methods, mRMR dominates regardless of the number of features.
Overall, only RF, mRMR and LassoNet result in best $p$ and best $2p$ scores $> 50\%$ when $m > 256$. All remaining methods except LassoNet appear to perform close to random for $m \ge 256$. LassoNet starts performing random for $m \ge 1024$.

\subsection{\ringxorsum \ is challenging for all methods}
\begin{figure}
    \includegraphics[width=\textwidth]{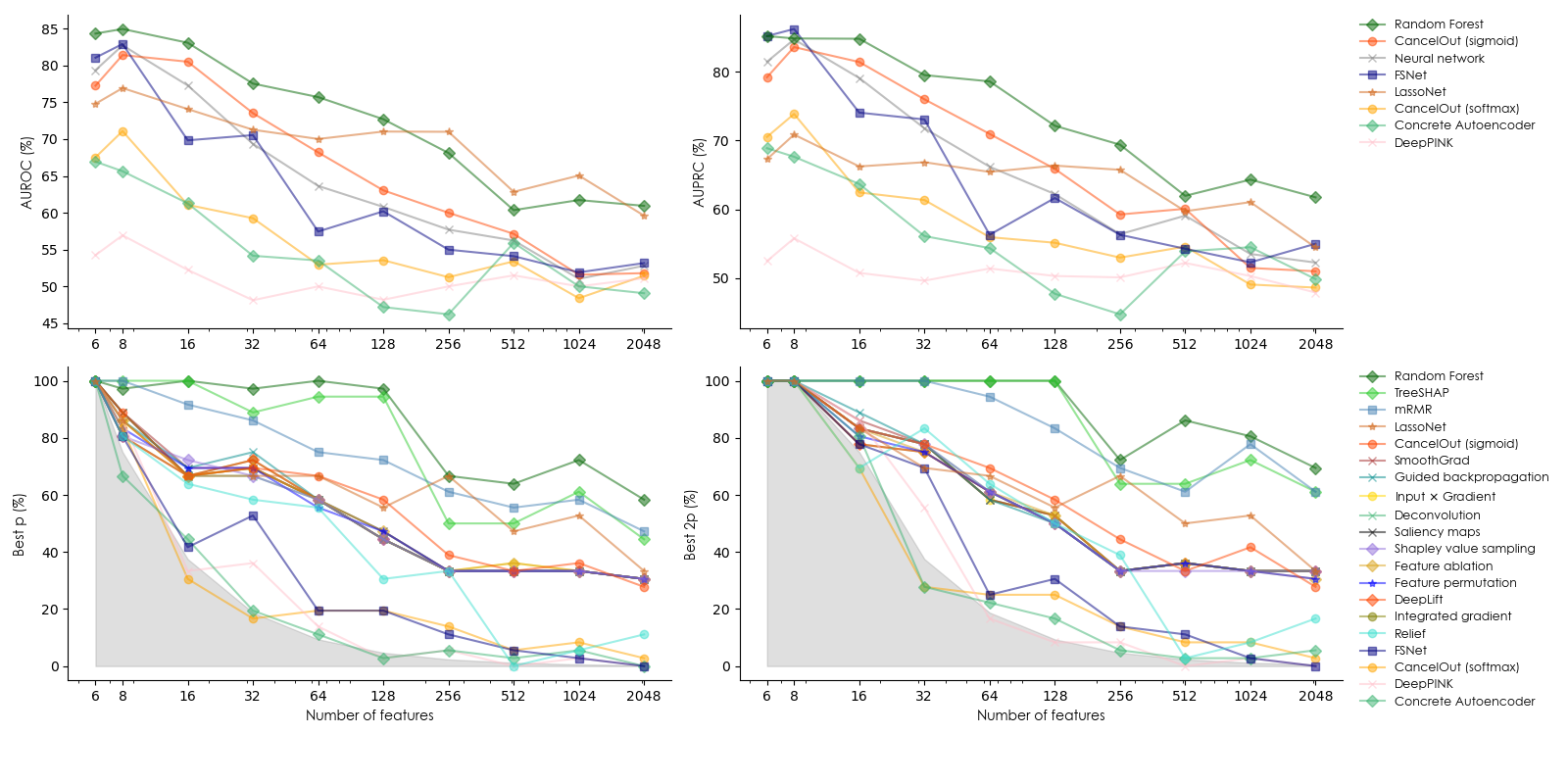}
    \caption{Performance of the different models and feature selection methods on the \ringxorsum \ dataset. (Top) AUROC and AUPRC of each trained model as a function of the number of features. (Bottom) Percentage of relevant features selected by each FS method in the top \(p\) and \(2p\), respectively. Shaded areas correspond to the best $p$ and best $2p$ scores of a dummy FS method that performs worse than random. Methods have been sorted by decreasing order of average performance in the legend.}
    \label{fig:benchmark-ring-xor-sum}
\end{figure}

There are six relevant features in this dataset, two of them being linear with some additive noise. The prediction problem appears to be quite difficult, with AUROC and AUPRC values $<$ 0.85 even for \(p = 6\) when no decoy feature is added (see Fig. \ref{fig:benchmark-ring-xor-sum}).
Contrary to the previous experiments, overall performance drops gradually with $m$.
This can be explained by the slightly larger number of relevant features ($p = 6$) as well as their heterogeneity, allowing models to detect the features that are the easiest to pick, given the characteristics of each model (\xor \ features for NNs, \ring \ features for mRMR and tree-based models, SUM features for most models).
RF/TreeSHAP and mRMR are the best FS methods, with the best $p$ scores $\ge 40\%$ and the best $2p$ scores $\ge 60\%$ for each value of $m$. Concrete autoencoder ranks last as FS method, and DeepPINK ranks last as a predictive model.
Finally, we observe that CancelOut Sigmoid clearly outperforms its softmax variant by a large margin.

\subsection{Disentangling causal from spurious effects on DAG is too challenging for FS methods}
In the fifth and last dataset in this benchmark, which has been generated by a graphical model (see Methods), relevant features can be defined either as features that are causal for the observed variable $Y$, or features that are expected to correlate with $Y$ due to indirect (confounding) effects. We refer to this dataset as DAG.
In Table \ref{tab:benchmark-dag}, we reported the best $p$ and $2p$ scores based on these two definitions.
We can see that, in both settings, TreeSHAP and its underlying Random Forest model are outperforming other methods. Concrete autoencoder, FSNet, CancelOut (softmax) and DeepPINK fail at detecting relevant features. Among feature attribution methods, Input $\times$ Gradient, Feature permutation and Shapley value sampling slightly improve over the other methods. Overall, none of the benchmarked approaches seems to be relatively better at disentangling indirect correlations from causal effects on the DAG dataset.
\begin{table}
\centering
{\resizebox{\columnwidth}{!}{\begin{tabular}{lrrrrrr}\toprule
Dataset & \multicolumn{2}{c}{\% causal features} & \multicolumn{2}{c}{\% correlated features} & \multicolumn{2}{c}{Performance (\%)} \\
\cmidrule(lr){2-3} \cmidrule(lr){4-5} \cmidrule(lr){6-7}
Method & Best p & Best 2p & Best p & Best 2p & AUROC & AUPRC \\ \midrule
Saliency maps & 14.3 & 21.4 & 9.5 & 14.6 & - & - \\
Integrated gradient & 14.3 & 21.4 & 9.7 & 14.6 & - & - \\
DeepLift & 14.3 & 21.4 & 9.7 & 14.6 & - & - \\
Input $\times$ Gradient & 19.0 & 26.2 & 10.1 & 14.2 & - & - \\
SmoothGrad & 14.3 & 21.4 & 9.5 & 14.6 & - & - \\
Guided backpropagation & 14.3 & 21.4 & 9.5 & 14.6 & - & - \\
Deconvolution & 14.3 & 21.4 & 9.5 & 14.6 & - & - \\
Feature ablation & 19.0 & 26.2 & 9.7 & 14.2 & - & - \\
Feature permutation & 16.7 & 21.4 & 10.3 & 14.4 & - & - \\
Shapley value sampling & 16.7 & 26.2 & 10.3 & 14.6 & - & - \\
\midrule
mRMR & 16.7 & 19.0 & 6.6 & 11.5 & - & - \\
LassoNet & 14.3 & 14.3 & 6.8 & 11.1 & 71.9 & 66.0 \\
Relief & 14.3 & 14.3 & 6.8 & 9.5 & - & - \\
Concrete Autoencoder & 2.4 & 11.9 & 4.9 & 8.0 & 50.1 & 52.8 \\
FSNet & 2.4 & 4.8 & 3.5 & 7.2 & 56.7 & 55.5 \\
CancelOut (softmax) & 2.4 & 4.8 & 5.8 & 8.2 & 46.5 & 48.5 \\
CancelOut (sigmoid) & 16.7 & 21.4 & 9.7 & 13.8 & 56.2 & 56.8 \\
DeepPINK & 0.0 & 0.0 & 4.3 & 8.4 & 50.0 & 50.0 \\
Random Forest & 21.4 & 40.5 & 12.8 & \textbf{17.9} & \textbf{75.4} & \textbf{73.9} \\
TreeSHAP & \textbf{28.6} & \textbf{42.9} & \textbf{13.4} & 17.3 & - & - \\
\bottomrule
\end{tabular}}}{}
\vspace*{3mm}
\caption{Best p and best 2p scores of all feature selection methods on the DAG dataset. $p=7$ when considering only causal features as relevant (first multi-column) and $p=81$ when also including confouders (second multi-column). Additionally, AUROC and AUPRC scores have been reported for embedded methods.\label{tab:benchmark-dag}}
\end{table}

\subsection{Benchmark summary}

Complementary to the presented figures, we summarised in Tab. \ref{tab:benchmark} the best $p$ and best $2p$ scores for all FS methods on each dataset. For readability purposes, we only report the mean scores, computed across all values of $m$. The table is organised in two parts: instance-level feature attribution methods (\emph{a posteriori} gradient-based methods such as Saliency Maps), and the remaining approaches. Best performing methods from each category are highlighted in bold. Among model-based and \textit{a priori} FS techniques, TreeSHAP and its underlying model Random Forests both outperform all approaches by a large margin on all datasets but \xor. Overall, mRMR appears to be the second best-performing technique, despite its underperformance on \xor. Complementary to mRMR, LassoNet achieved maximal performance on \xor, while getting average results on the remaining datasets.

Instance-level \textit{a posteriori} methods do not show significant differences, except on the DAG dataset. In the latter setting, Input $\times$ Gradient, Feature Ablation and Shapley value sampling seem to perform relatively better.

In order to compare the methods in situations where the curse of dimensionality is exacerbated, we reported the same results on sub-sampled versions of the same datasets, with $n \in \{250, 500\}$. These results are shown in Supp. Tables 1 and 2. We observed that mRMR, tree-based methods and LassoNet consistently outperform other methods in the same settings.

\begin{table}
\centering
{\resizebox{\columnwidth}{!}{\begin{tabular}{lrrrrrrrrrr}\toprule
Dataset & \multicolumn{2}{c}{\ring} & \multicolumn{2}{c}{\xor} & \multicolumn{2}{c}{\ringxor} & \multicolumn{2}{c}{\ringxorsum} & \multicolumn{2}{c}{DAG} \\
\cmidrule(lr){2-3} \cmidrule(lr){4-5} \cmidrule(lr){6-7} \cmidrule(lr){8-9} \cmidrule(lr){10-11}
Method & Best k & Best 2k & Best k & Best 2k & Best k & Best 2k & Best k & Best 2k & Best k & Best 2k \\ \midrule
Saliency maps & 25.8 & 34.8 & 53.8 & 54.5 & 35.0 & 42.5 & 56.1 & 60.8 & 14.3 & 21.4 \\
Integrated gradient & 24.2 & 36.4 & 53.0 & 54.5 & 35.4 & 41.7 & 55.3 & 60.0 & 14.3 & 21.4 \\
DeepLift & 24.2 & 36.4 & 53.0 & 55.3 & 34.6 & 42.5 & 55.3 & 60.0 & 14.3 & 21.4 \\
Input $\times$ Gradient & 26.5 & 34.8 & 54.5 & 54.5 & 35.8 & 43.3 & 55.8 & 60.8 & 19.0 & 26.2 \\
SmoothGrad & 25.8 & 36.4 & 53.8 & 54.5 & 36.2 & 42.9 & 56.1 & 61.1 & 14.3 & 21.4 \\
Guided backpropagation & 25.8 & 36.4 & 53.8 & 54.5 & 35.4 & 42.5 & 56.4 & 61.1 & 14.3 & 21.4 \\
Deconvolution & 25.8 & 35.6 & 53.8 & 55.3 & 35.0 & 43.3 & 55.6 & 60.8 & 14.3 & 21.4 \\
Feature ablation & 25.0 & 34.8 & 54.5 & 54.5 & 34.2 & 42.1 & 56.1 & 60.6 & 19.0 & 26.2 \\
Feature permutation & 25.8 & 35.6 & 53.0 & 54.5 & 32.9 & 41.7 & 55.6 & 60.0 & 16.7 & 21.4 \\
Shapley value sampling & 23.5 & 37.9 & 52.3 & 53.8 & 35.4 & 42.9 & 55.3 & 60.6 & 16.7 & 26.2 \\
\midrule
mRMR & \textbf{100.0} & \textbf{100.0} & 12.5 & 28.3 & 81.7 & 88.8 & 74.7 & 84.7 & 16.7 & 19.0 \\
LassoNet & 34.8 & 35.6 & \textbf{81.8} & \textbf{81.8} & 44.6 & 52.9 & 64.2 & 67.8 & 14.3 & 14.3 \\
Relief & 40.2 & 45.5 & 72.7 & 74.2 & 37.1 & 42.1 & 43.9 & 53.3 & 14.3 & 14.3 \\
Concrete Autoencoder & 19.7 & 24.2 & 22.7 & 27.3 & 19.2 & 30.0 & 25.8 & 36.4 & 2.4 & 11.9 \\
FSNet & 21.2 & 28.0 & 31.1 & 38.6 & 25.4 & 33.3 & 33.3 & 43.1 & 2.4 & 4.8 \\
CancelOut (softmax) & 14.4 & 25.0 & 22.0 & 28.8 & 18.8 & 29.2 & 30.0 & 38.1 & 2.4 & 4.8 \\
CancelOut (sigmoid) & 21.2 & 31.1 & 67.4 & 72.0 & 35.8 & 46.7 & 58.6 & 63.6 & 16.7 & 21.4 \\
DeepPINK & 17.4 & 25.0 & 17.4 & 24.2 & 20.4 & 32.9 & 28.1 & 37.8 & 0.0 & 0.0 \\
Random Forest & \textbf{100.0} & \textbf{100.0} & 54.5 & 59.8 & \textbf{88.8} & \textbf{95.4} & \textbf{85.3} & \textbf{90.8} & 21.4 & 40.5 \\
TreeSHAP & \textbf{100.0} & \textbf{100.0} & 40.2 & 43.9 & 81.7 & 88.3 & 78.3 & 86.1 & \textbf{28.6} & \textbf{42.9} \\
\bottomrule
\end{tabular}}}{}
\vspace*{3mm}
\caption{Best k and best 2k score percentages on the 5 datasets. For the first 4 datasets, scores have been averaged over $m \in \{2, 4, 6, 8, 16, 32, 64, 128, 256, 512, 1024, 2048\}$.Top and bottom parts of the table correspond to instance-level feature attribution and embedded/filter FS methods, respectively. Best performing methods are highlighted in bold.\label{tab:benchmark}}
\end{table}

\subsection{Bootstrapping influences the quality of feature attribution scores}
\label{attr-train-test}

Since practitioners may use \textit{a posteriori} SM methods in different ways, and since it was unclear whether using points from the training set can indeed improve the quality of feature ranking, we investigated whether the latter can benefit from bootstrapping and the use of the training set. In the first setting, we computed one feature importance vector per instance from the held-out set after training, and averaged them across the whole held-out set. In a second stage, we added a bootstrapping component by repeating this step 10 times and re-training the model each time on a random sample composed of 80\% of the training set (sampling with replacement). The final importance vectors have been obtained by averaging across the 10 runs. In the third setting, we removed bootstrapping but computed the feature importance vectors on the points from the training set only, and left the held-out set unused.

\begin{figure}
    \includegraphics[width=\textwidth]{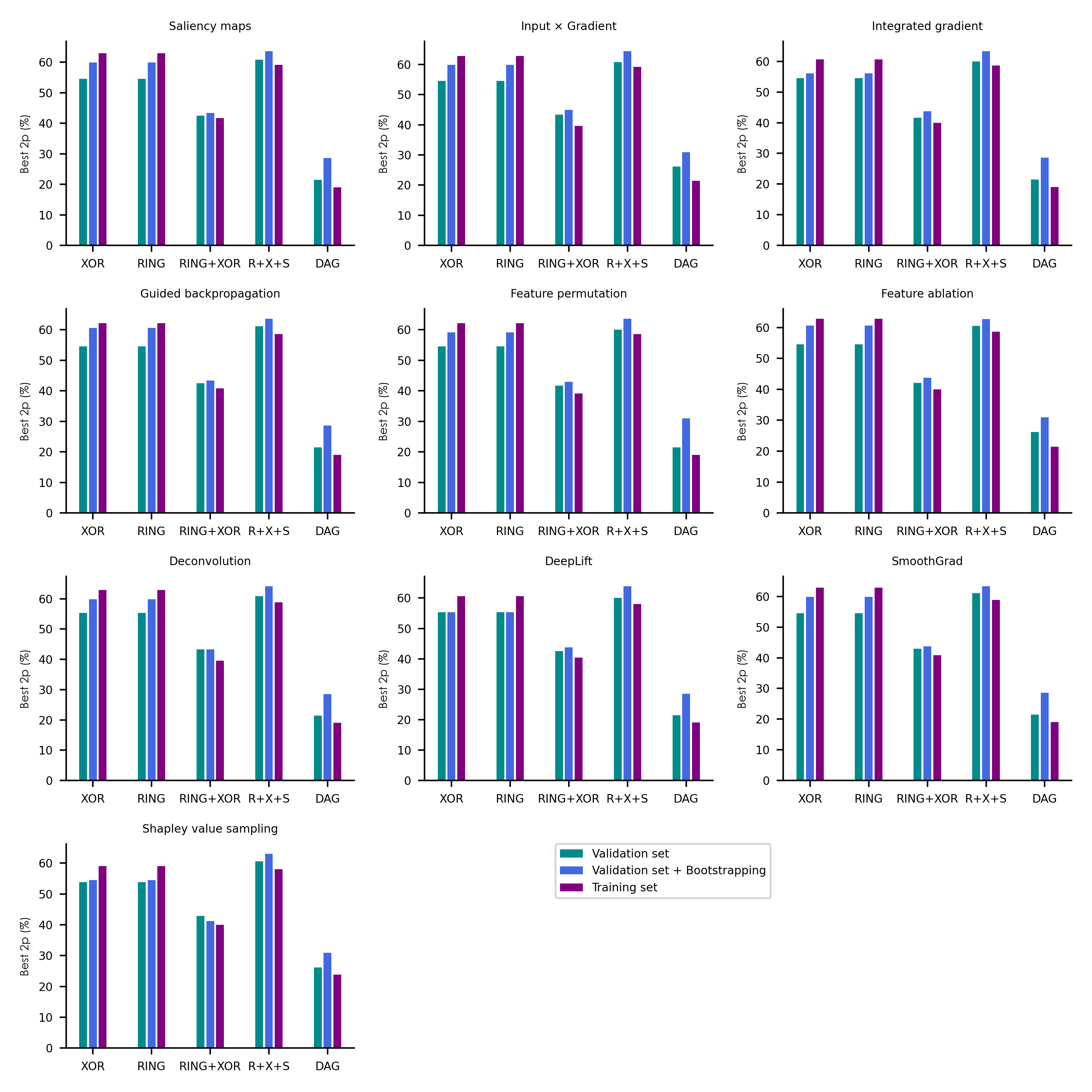}
    \caption{Average best $2p$ score on each of the 5 datasets, using three different approaches to infer feature importances from instance-level feature attribution methods. In the first setting, only the points from the held-out set have been used to rank features. In the second case, bootstrapping was performed by randomly sampling (with replacement) 80\% of the training set before training the NN. In the last setting, only the points from the training set have been used.}
    \label{fig:bootstrapping}
\end{figure}

\section{Discussion}

\subsection{Feature selection and modeling quality are interdependent}

Care should be taken in interpreting the results presented in our study. Indeed, each FS method necessarily relies on some modeling assumptions, and the relevance of selected features is heavily impacted by the model's adequacy to the data. In particular, the inference process does not guarantee to yield the optimal solution (the set of parameters that produces the least generalisation error). Despite the flexibility of NNs, as described by the universal approximation theorem \cite{cybenko1989approximation} for example, the optimal architecture choice (with smallest generalisation error) is unknown and can only be found by cross-validation. The eventual lack of regularisation, coupled with the presence of a large number of input features, therefore increasing the effective number of parameters, is likely to drive the model towards learning from irrelevant correlations.
In such case, because the model's decision process relies on irrelevant features, the FS method building on this model will necessarily attach higher importance to those features. Therefore, FS techniques based on NNs might be hard to exploit in practice, as they require building the most accurate model, and guiding the inference process to the optimal solution. This goal is easier to reach when provided with sufficiently deep insights about the data. In this sense, FS resembles a chicken-and-egg problem, and because of that FS methods cannot be blindly applied on a dataset without minimal prior knowledge of the data and the limitations of the model used.

In particular, all feature attribution / SM methods considered in this study, as well as many of the other FS techniques, rely on a neural network that requires proper training. Because the number of parameters varied with the number of input features, it remains highly probable that the corresponding models either under- or overfitted the data in some situations, regardless of the presence of dropout and L2 regularisation. Overall, there is no guarantee that the Adam optimizer consistently guided the model to the globally-optimal solution, or that the generalisation error was minimal.

\subsection{Tree-based modeling and decision tree induction are two separate concepts}

Random forests have largely outperformed other methods on the \ring, \ringxor, \ringxorsum \ and DAG datasets, both as a FS technique and as predictive models. However, the dataset where RFs perform comparatively (to other methods) worse is \xor, which consists of data points obeying a simple logical rule. Such rule can be perfectly captured through a piece-wise linear function. In particular, among all off-the-shelf ML models, decision trees are the optimal choice for modeling such data, as only three decision splits should be theoretically sufficient for perfect segregation of the classes. However, performance of RFs is sub-optimal as observed in Fig. \ref{fig:benchmark-xor}.

The quality of its feature selection and prediction is not linked to the sophistication of the modeling \textit{per se}, \textit{but rather the optimality of the decision tree induction algorithm}. Indeed, Scikit-learn's implementation is based on the heuristic CART algorithm \cite{breiman2017classification}, which is unlikely to infer the tree with highest information gain and minimal number of nodes. In particular in the \xor \ dataset, selecting one of the two relevant features as first decision split is not sufficient, as it does not produce any change in the class proportions within the newly obtained hyper-paralelipipeds. Therefore, RFs are counter-intuitively better at growing from ring-shaped data, since any split on one of the 2 corresponding features results in an increase of class purity (strictly positive information gain). On the XOR dataset, optimal inference would require a lookahead of 1 feature, or bivariate decision splits. In conclusion, the relatively lower performance of RFs on the \xor \ dataset can mostly be attributed to the sub-optimality of its underlying tree induction algorithm.

\subsection{Univariate feature selection remains relevant in a high dimensionality context}

Although the datasets have been constructed in a way that they are highly challenging for both linear and univariate FS methods, it must be noted that (nonlinear) univariate filter methods remain highly relevant in some contexts. First, they are computationally more efficient (and embarrassingly parallel), making them competitive in a high-dimensional setting (e.g. Whole Genome Sequencing data). Second, they are still capable of capturing relevant features when they \emph{individually} correlate with the explained variable, in a nonlinear fashion. This is shown by the maximal performance ($100\%$ best $p$ score for any value of $m$) of mutual information (MI) on the \ring \ dataset (as shown in Suppl. Tab. 3), suggesting that mutual information (MI) consistently detects the reduction of entropy caused by the ring-shaped function that generated the data. 
Indeed, this ring-shaped data is leaking information about the class at the level of individual features (the probability distribution of each feature is altered, when conditioned on the class distribution). Because real-life problems are more likely to exhibit univariate nonlinear correlations than our artificial datasets, simple information-theoretic approaches could remain highly relevant. However, data availability is a crucial prerequisite for accurate estimation of MI.


\section{Conclusion}
In this paper, we investigated the usefulness of nonlinear FS approaches in the case of high-dimensional data with a low sample-to-feature ratio.
What emerges from the results is that Random Forests and Relief outperform neural network-based approaches on all datasets that exhibit nonlinear correlations between individual features and the target variable (all but the \xor \ dataset).
Therefore, DNN models might not be the best choice for feature selection on datasets with these characteristics, i.e. a low density of relevant features, a prevalence of nonlinear patterns and variance homogeneity (for both predictive and decoy features).
In real-life applications, we mainly encounter that relationships among the features in the data tend to be additive. However, our study indicates that when both additive and nonlinearly entangled features are present, we can only see the former unless a considerable number of samples are available compared to the input feature size.

\bibliographystyle{unsrt}

\bibliography{main}

\end{document}



\section{Supplementary materials}

\subsection{Description of the DAG dataset}

The DAG dataset contains $n=1000$ observations and $m=2000$ features, where each feature $X_i$ is defined as follows:
\begin{align*}
    X_i & = h\left(\alpha_i Z_i + \beta_i\right) \\
    Z_i & = \epsilon_i + \frac{1}{\vert p(i)\vert}\sum_{j \in p(i)} \gamma_{ji}X_j \\
    \epsilon_i & \sim \mathcal{N}(0, \sigma)
\end{align*}
$h$ is a nonlinear function, $\alpha_i, \beta_i$ are scaling and centering parameters used to standardize the inputs provided to $h$, and $p(i)$ is a set containing the parents of node $i$ in the graphical model.

We used the Sigmoid activation function to implement the nonlinearity $h$. $\alpha_i$ is computed as the inverse of the standard deviation of the $n$ $X_i$ values, and $\beta_i$ the negative of their mean. Each weight $\gamma_{ji}$ was drawn from a uniform distribution $\mathcal{U}(-1, 1)$ beforehand. $0,3$ was set as the default value for the standard deviation of the noise term.

The adjacency matrix of the underlying graphical model was constructed as follows:
\begin{itemize}
    \item We started from a null graph with $m$ nodes.
    \item Each edge $X_i \rightarrow X_j$ was added with probability 0,5\%.
    \item The lower triangle of the adjacency matrix was enforced to be $0$, in order to ensure the acyclicity of the graph.
    \item Based on the desired number of irrelevant features (we used $1000$), we zeroed randomly selected rows/columns.
    \item Based on the desired number of relevant causal features (we used $20$), we randomly added $X_i \rightarrow Y$ edges.
\end{itemize}

Finally, we turned $Y$ into a binary variable by using the median of the $n$ values as a cutoff. In particular, $Y = 1$ when the corresponding value exceeds the median.

\subsection{DeepPINK FS method: how the knockoff features are defined}

\subsubsection{DAG dataset}

The definition of Model-X knockoffs features is the following.

\begin{Def}\label{def:knockoff}
Model-X knockoffs features for the family of random variables $\bm{x} = (X_1,...,X_p)^T$ are a new family of random variables  $\bm{\widetilde{x}} = (\widetilde{X}_1,...,\widetilde{X}_p)^T$ that satifies the following properties:
\begin{itemize}
\item[(i)] $(\bm{x},\bm{\widetilde{x}})_{\mathrm{swap(S)}} \overset{d}{=} (\bm{x},\bm{\widetilde{x}})$ for each subset $\mathrm{S} \in {1,...,p}$
\item[(ii)] $\bm{\widetilde{x}} \bigCI Y\mid\bf{x}$
\end{itemize}
Where $\mathrm{swap(S)}$ means swapping the entries $X_{j}$ and $\widetilde{X}_{j}$ for each $j \in S$ and $\overset{d}{=}$ denotes equal in distribution. $Y$ is the response of feature vector $\bm{x}$.
\end{Def}
Assuming  $\bf{x} \sim \mathcal{N}(0,\,\Sigma)$ with  $\Sigma \in \mathbb{R}^{p \times p }$, we can build the knockoff features as follows:

\begin{align}
    \bm{\widetilde{x}} \vert \bm{x} \sim  \mathcal{N}(\bm{x} - D \Sigma^{-1}\bm{x},2 D - D \Sigma^{-1} D)
\end{align}
Where $D$ is a diagonal matrix with all its diagonal components being positive.
In our work we computed the knockoff features by applying this definition. Let's note that condition \textit{(i)} of Definition \ref{def:knockoff} holds up to the moments of second-order.

\subsection{Feature attribution methods rely on different interpretation mechanisms}

\begin{figure}[H]
    \includegraphics[width=\textwidth]{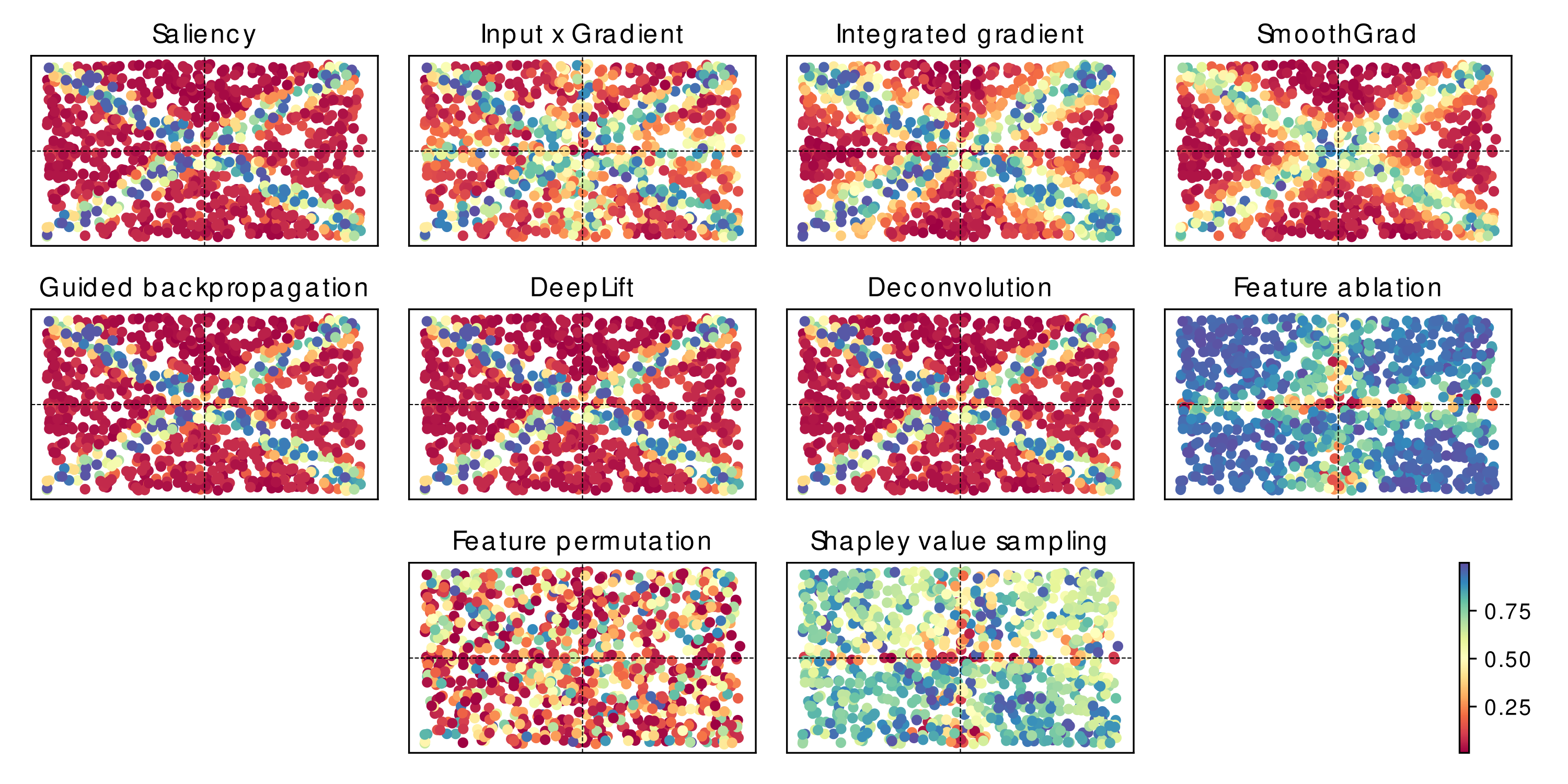}
    \caption{Visualisation of the attributions made by different methods on the two first features of the \xor \ dataset. The color of each input vector is computed as the minimum of importances from features 1 and 2 (in absolute terms) divided by the maximum of importances from features 1 and 2 (in absolute terms). In particular, blue points are treated as being of equal importance. Dashed lines represent the optimal decision boundaries for such dataset.}
    \label{fig:interpretation-xor}
\end{figure}

In Fig. \ref{fig:interpretation-xor}, we reported the feature importances provided by all the feature attribution methods from our benchmark. Feature importances were computed for each input vector from the test set (of each fold) within the \xor \ dataset, for $m = 2$. The colour was assigned according to the value of $\frac{min(|\zeta_1|,|\zeta_2|)}{max(|\zeta_1|,|\zeta_2|)}$, where  $\zeta_1$ and $\zeta_2$ are the importances of features 1 (x-axis) and 2 (y-axis), respectively. 

It must be noted that neural networks systematically reached 100\% of AUROC and AUPRC on this dataset when $m=2$, suggesting that their decision boundaries closely followed the dashed lines shown in Fig. \ref{fig:interpretation-xor}. Therefore, it is surprising to observe significant differences between some of the different feature attribution techniques.
Other gradient-based approaches, like Guided Backpropagation or Deconvolution produced similar results. 
The Input $\times$ gradient method shows a different pattern due to the multiplication with the input features (a disruption of the vanilla saliency pattern occurs near the decision boundaries). 
Ratios yielded by Feature Permutation appeared to be extremely noisy compared to all other techniques, which can be attributed to its natural inability of providing instance-level feature importances.

Feature Ablation and Shapley Value Sampling provided different insights about the data. It appears that these methods delineate better the decision boundaries. 
However, these interpretability differences do not privilege any FA methods. The FS needed for our test is global and relies on the averaging of FA vectors across all points in the dataset, which will necessarily capture the overall importance of the two features.
A similar analysis was conducted on the \ring \ dataset, and provided in Suppl. Fig. \suppfiginterpretation.

\begin{figure}[H]
    \centering
    \includegraphics[width=\textwidth]{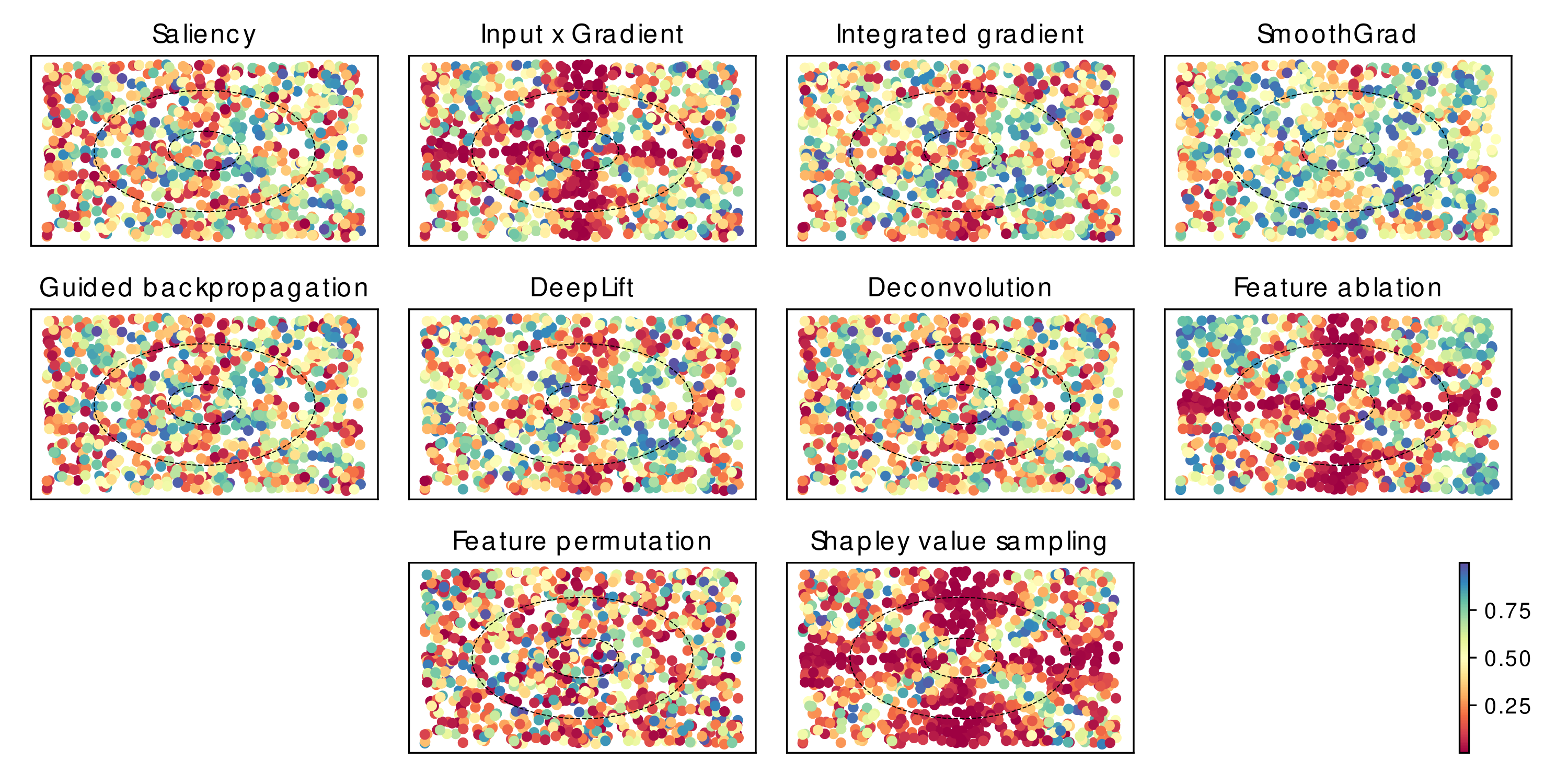}
    \caption{Visualisation of the attributions made by different methods on the two first features of the \ring \ dataset. The color of each input vector is computed as the minimum of importances from features 1 and 2 (in absolute terms) divided by the maximum of importances from features 1 and 2 (in absolute terms). In particular, blue points are treated as being of equal importance. Dashed lines represent the optimal decision boundaries for such dataset.}
    \label{fig:interpretation-ring}
\end{figure}

\subsection{Exacerbating the curse of dimensionality by decreasing the sample size}

\begin{table}[H]
\centering
\label{tab:benchmark}{\resizebox{\columnwidth}{!}{\begin{tabular}{lrrrrrrrrrr}\toprule
Dataset & \multicolumn{2}{c}{\ring} & \multicolumn{2}{c}{\xor} & \multicolumn{2}{c}{\ringxor} & \multicolumn{2}{c}{\ringxorsum} & \multicolumn{2}{c}{DAG} \\
\cmidrule(lr){2-3} \cmidrule(lr){4-5} \cmidrule(lr){6-7} \cmidrule(lr){8-9} \cmidrule(lr){10-11}
Method & Best p & Best 2p & Best p & Best 2p & Best p & Best 2p & Best p & Best 2p & Best p & Best 2p \\ \midrule
Saliency maps & 19.7 & 25.0 & 38.6 & 44.7 & 20.8 & 32.5 & 40.0 & 48.3 & 16.7 & 21.4 \\
Integrated gradient & 17.4 & 25.0 & 37.1 & 40.2 & 20.8 & 33.8 & 40.6 & 48.3 & 16.7 & 21.4 \\
DeepLift & 16.7 & 25.0 & 37.9 & 40.2 & 21.2 & 33.8 & 40.8 & 49.2 & 16.7 & 21.4 \\
Input $\times$ Gradient & 19.7 & 25.8 & 38.6 & 41.7 & 22.5 & 32.1 & 39.4 & 48.1 & 16.7 & 19.0 \\
SmoothGrad & 18.2 & 25.0 & 39.4 & 43.2 & 21.7 & 34.2 & 39.7 & 47.8 & 16.7 & 21.4 \\
Guided backpropagation & 19.7 & 25.8 & 38.6 & 40.9 & 21.2 & 32.5 & 40.3 & 49.2 & 16.7 & 21.4 \\
Deconvolution & 18.2 & 26.5 & 38.6 & 42.4 & 20.8 & 33.3 & 39.7 & 49.2 & 16.7 & 21.4 \\
Feature ablation & 18.9 & 26.5 & 37.9 & 40.2 & 20.8 & 32.9 & 38.6 & 47.8 & 16.7 & 19.0 \\
Feature permutation & 18.2 & 25.8 & 37.1 & 40.2 & 21.7 & 32.1 & 38.9 & 49.2 & 14.3 & 19.0 \\
Shapley value sampling & 18.2 & 25.0 & 34.8 & 37.1 & 21.2 & 33.3 & 41.1 & 48.3 & 16.7 & 16.7 \\
\midrule
mRMR & 71.2 & 81.1 & 15.2 & 25.0 & 30.4 & 39.2 & 31.1 & 44.2 & 14.3 & 19.0 \\
LassoNet & 17.4 & 28.0 & \textbf{63.6} & \textbf{63.6} & 31.2 & 35.4 & 45.3 & 58.9 & 14.3 & 14.3 \\
Relief & 27.3 & 35.6 & 53.8 & 56.1 & 25.8 & 33.3 & 33.1 & 44.2 & 14.3 & 16.7 \\
Concrete Autoencoder & 18.9 & 27.3 & 15.2 & 25.8 & 21.2 & 32.9 & 24.7 & 35.0 & 9.5 & 11.9 \\
FSNet & 19.7 & 25.0 & 25.0 & 32.6 & 23.3 & 32.5 & 29.4 & 41.4 & 0.0 & 0.0 \\
CancelOut (softmax) & 15.2 & 22.0 & 28.0 & 34.1 & 21.2 & 28.7 & 30.3 & 38.6 & 14.3 & 16.7 \\
CancelOut (sigmoid) & 21.2 & 25.0 & 50.8 & 54.5 & 24.6 & 33.3 & 41.1 & 51.1 & 14.3 & 14.3 \\
DeepPINK & 17.4 & 28.0 & 18.2 & 25.8 & 21.2 & 31.7 & 26.9 & 37.8 & 0.0 & 0.0 \\
Random Forest & \textbf{87.9} & \textbf{95.5} & 47.7 & 50.8 & \textbf{38.3} & \textbf{50.0} & \textbf{51.4} & \textbf{62.5} & \textbf{23.8} & 35.7 \\
TreeSHAP & 68.2 & 88.6 & 37.9 & 47.0 & 30.4 & 41.2 & 47.8 & 57.2 & \textbf{23.8} & \textbf{40.5} \\
\bottomrule
\end{tabular}}}{}
\caption{Best p and best 2p score percentages on the 5 datasets, for sample size $n=250$. For the first 4 datasets, scores have been averaged over $m \in \{2, 4, 6, 8, 16, 32, 64, 128, 256, 512, 1024, 2048\}$.Top and bottom parts of the table correspond to instance-level feature attribution and embedded/filter FS methods, respectively. Best performing methods are highlighted in bold.}
\end{table}

\begin{table}[H]
\centering
\label{tab:benchmark}{\resizebox{\columnwidth}{!}{\begin{tabular}{lrrrrrrrrrr}\toprule
Dataset & \multicolumn{2}{c}{\ring} & \multicolumn{2}{c}{\xor} & \multicolumn{2}{c}{\ringxor} & \multicolumn{2}{c}{\ringxorsum} & \multicolumn{2}{c}{DAG} \\
\cmidrule(lr){2-3} \cmidrule(lr){4-5} \cmidrule(lr){6-7} \cmidrule(lr){8-9} \cmidrule(lr){10-11}
Method & Best p & Best 2p & Best p & Best 2p & Best p & Best 2p & Best p & Best 2p & Best p & Best 2p \\ \midrule
Saliency maps & 25.8 & 34.8 & 48.5 & 49.2 & 26.7 & 39.2 & 43.6 & 51.1 & 28.6 & 31.0 \\
Integrated gradient & 24.2 & 32.6 & 47.0 & 50.0 & 28.7 & 37.5 & 42.8 & 50.8 & 28.6 & 31.0 \\
DeepLift & 25.0 & 33.3 & 47.7 & 49.2 & 27.5 & 37.5 & 42.2 & 51.4 & 28.6 & 31.0 \\
Input $\times$ Gradient & 25.0 & 35.6 & 46.2 & 47.7 & 28.3 & 38.3 & 42.8 & 51.1 & 23.8 & 28.6 \\
SmoothGrad & 25.8 & 34.8 & 48.5 & 49.2 & 28.7 & 38.3 & 44.4 & 51.4 & 28.6 & 31.0 \\
Guided backpropagation & 25.8 & 35.6 & 49.2 & 50.0 & 27.9 & 38.8 & 43.3 & 51.9 & 28.6 & 31.0 \\
Deconvolution & 27.3 & 35.6 & 48.5 & 50.8 & 27.9 & 39.2 & 43.3 & 51.9 & 28.6 & 31.0 \\
Feature ablation & 26.5 & 32.6 & 48.5 & 50.0 & 26.7 & 37.5 & 42.2 & 50.6 & 23.8 & 28.6 \\
Feature permutation & 25.0 & 36.4 & 48.5 & 50.0 & 27.1 & 37.1 & 43.9 & 51.7 & 26.2 & 28.6 \\
Shapley value sampling & 24.2 & 34.8 & 43.9 & 45.5 & 26.7 & 36.2 & 42.2 & 51.7 & 21.4 & 31.0 \\
\midrule
mRMR & \textbf{98.5} & \textbf{100.0} & 16.7 & 25.0 & 44.2 & 59.2 & 44.4 & 58.9 & 19.0 & 21.4 \\
LassoNet & 25.8 & 30.3 & \textbf{72.7} & \textbf{72.7} & 38.3 & 43.8 & 58.1 & 63.1 & 14.3 & 14.3 \\
Relief & 35.6 & 40.2 & 61.4 & 65.2 & 32.5 & 40.0 & 36.4 & 46.9 & 14.3 & 14.3 \\
Concrete Autoencoder & 19.7 & 25.8 & 16.7 & 27.3 & 22.5 & 31.7 & 25.0 & 32.5 & 2.4 & 11.9 \\
FSNet & 18.2 & 24.2 & 33.3 & 40.2 & 24.2 & 32.5 & 43.1 & 51.4 & 0.0 & 2.4 \\
CancelOut (softmax) & 22.7 & 31.1 & 25.8 & 31.8 & 22.9 & 33.8 & 33.6 & 42.5 & 9.5 & 9.5 \\
CancelOut (sigmoid) & 26.5 & 31.8 & 53.8 & 56.8 & 30.4 & 38.8 & 51.1 & 60.0 & 16.7 & 21.4 \\
DeepPINK & 17.4 & 24.2 & 17.4 & 24.2 & 22.1 & 28.7 & 29.2 & 38.9 & 4.8 & 4.8 \\
Random Forest & 97.7 & 99.2 & 50.0 & 55.3 & \textbf{65.8} & \textbf{75.4} & \textbf{64.4} & \textbf{71.1} & 28.6 & 40.5 \\
TreeSHAP & 90.2 & 97.0 & 35.6 & 40.2 & 49.6 & 63.7 & 62.5 & 69.4 & \textbf{33.3} & \textbf{42.9} \\
\bottomrule
\end{tabular}}}{}
\caption{Best p and best 2p score percentages on the 5 datasets, for sample size $n = 500$. For the first 4 datasets, scores have been averaged over $m \in \{2, 4, 6, 8, 16, 32, 64, 128, 256, 512, 1024, 2048\}$.Top and bottom parts of the table correspond to instance-level feature attribution and embedded/filter FS methods, respectively. Best performing methods are highlighted in bold.}
\end{table}

\subsection{FS accuracy of univariate feature selection based on mutual information}

\begin{table}[H]
\centering
{\resizebox{\columnwidth}{!}{\begin{tabular}{lrrrrrrrrrr}\toprule
 & \multicolumn{2}{c}{\ring} & \multicolumn{2}{c}{\xor} & \multicolumn{2}{c}{\ringxor} & \multicolumn{2}{c}{\ringxorsum} & \multicolumn{2}{c}{DAG} \\
\cmidrule(lr){2-3} \cmidrule(lr){4-5} \cmidrule(lr){6-7} \cmidrule(lr){8-9} \cmidrule(lr){10-11}
$m$ & Best k & Best 2k & Best k & Best 2k & Best k & Best 2k & Best k & Best 2k & Best k & Best 2k \\ \midrule
8 & 100.0 & 100.0 & 66.7 & 75.0 & 87.5 & 100.0 & 94.4 & 100.0 & - & - \\
16 & 100.0 & 100.0 & 16.7 & 25.0 & 33.3 & 70.8 & 69.4 & 91.7 & - & -  \\
32 & 100.0 & 100.0 & 16.7 & 33.3 & 33.3 & 70.8 & 50.0 & 69.4 & - & -  \\
64 & 100.0 & 100.0 & 0.0 & 0.0 & 50.0 & 58.3 & 55.6 & 58.3 & - & -  \\
128 & 100.0 & 100.0 & 0.0 & 0.0 & 41.7 & 54.2 & 47.2 & 66.7 & - & -  \\
256 & 100.0 & 100.0 & 0.0 & 0.0 & 4.2 & 8.3 & 16.7 & 25.0 & - & -  \\
512 & 100.0 & 100.0 & 0.0 & 0.0 & 20.8 & 20.8 & 2.8 & 5.6 & - & -  \\
1024 & 100.0 & 100.0 & 0.0 & 0.0 & 8.3 & 16.7 & 8.3 & 13.9 & - & -  \\
2000 & - & - & - & - & - & - & - & - & 14.3 & 14.3  \\
2048 & 100.0 & 100.0 & 0.0 & 0.0 & 8.3 & 8.3 & 0.0 & 2.8 & - & -  \\
\bottomrule
\end{tabular}}}{}
\caption{Best k and best 2k score percentages of mutual information (MI) on the 5 datasets, for each value of $m \in \{2, 4, 6, 8, 16, 32, 64, 128, 256, 512, 1024, 2048\}$.}
\end{table}
